\documentclass[11pt]{article}

\usepackage[margin=0.88in]{geometry}
\usepackage[T1]{fontenc}
\usepackage[utf8]{inputenc}
\usepackage{lmodern}
\usepackage{microtype}
\usepackage{xspace}
\usepackage{amsmath,amssymb,amsthm,mathtools}
\usepackage{booktabs,tabularx,array,multirow}
\usepackage{graphicx}
\usepackage{algorithm}
\usepackage{algpseudocode}
\usepackage{enumitem}
\usepackage{xcolor}
\usepackage{natbib}
\usepackage{url}
\usepackage{hyperref}
\usepackage[nameinlink,capitalise,noabbrev]{cleveref}

\hypersetup{
  colorlinks=true,
  linkcolor=blue!55!black,
  citecolor=blue!55!black,
  urlcolor=blue!55!black,
  pdftitle={Drift-Aware LLM Routing with Sparse Contexts and Shared Budgets},
  pdfauthor={Cheung Hao Lee, Patrick Wong}
}

\newtheorem{theorem}{Theorem}
\newtheorem{lemma}{Lemma}
\newtheorem{proposition}{Proposition}
\newtheorem{corollary}{Corollary}
\newtheorem{assumption}{Assumption}
\theoremstyle{definition}

\theoremstyle{remark}
\newtheorem{remark}{Remark}
\crefname{assumption}{Assumption}{Assumptions}
\Crefname{assumption}{Assumption}{Assumptions}
\newcommand{\R}{\mathbb{R}}
\newcommand{\E}{\mathbb{E}}
\newcommand{\Prob}{\mathbb{P}}
\newcommand{\cA}{\mathcal{A}}
\newcommand{\cE}{\mathcal{E}}
\newcommand{\cF}{\mathcal{F}}
\newcommand{\cX}{\mathcal{X}}
\newcommand{\one}{\mathbf{1}}
\newcommand{\OPT}{\mathrm{OPT}}
\newcommand{\Reg}{\mathrm{Reg}}
\newcommand{\DRS}{\textnormal{\textsc{DR-Sparse}}\xspace}
\newcommand{\clip}{\operatorname{clip}}

\title{\textbf{Drift-Aware LLM Routing with Sparse Contexts}\\
\textbf{and Shared Budgets}}
\author{Cheung Hao Lee, Patrick Wong}
\date{}

\begin{document}
\maketitle

\begin{abstract}
A multi-model language service must route each request while preserving workload-level budgets for compute, latency, memory, or monetary cost. Two features make this problem materially harder than static model selection. Prompt representations are high dimensional, so only a small subset of embedding directions may predict the incremental value of a model; and both the request mix and the model frontier drift after launches, fine-tunes, quantization changes, and system updates. We formulate nonstationary sparse contextual routing with multiple knapsack constraints and an optional shadow-audit stream that evaluates a small fraction of prompts on several models.

We propose Drift-Aware Sparse Routing (\DRS). The policy estimates reward and resource use from a rolling audit window, routes using pessimistic reward and optimistic cost estimates, updates resource shadow prices online, and applies a hard meter before commitment. The analysis separates control from statistics. On any event with uniform prediction radii $\{\beta_t\}$, regret against a paced dynamic fluid benchmark is bounded by the sum of the radii, a capacity-buffer term, and an $O(\sqrt{T})$ pacing term. Under a sparse linear model and bounded drift $V_T$, rolling estimation gives
\[
\widetilde O\!\left(
T\sqrt{\frac{s}{\rho W}}+WV_T+\sqrt{T}
\right),
\]
where $s$ is sparsity, $\rho$ is the audit rate, and $W$ is the window length. Optimizing $W$ yields the usual stationary $\widetilde O(\sqrt{sT/\rho})$ rate when $V_T=0$ and a $\widetilde O(T^{2/3}(s/\rho)^{1/3}V_T^{1/3})$ adaptation term under drift.

A reproducible synthetic study with four model classes, two shared budgets, 28-dimensional contexts, and two model updates illustrates the mechanism. Rolling sparse routing retains roughly 97--98\% of the clairvoyant policy's utility and improves on a frozen router after the change points, while the resource meter remains feasible. The paper closes with a measurement protocol for real routing panels and a discussion of selective auditing, change detection, and integration with inference schedulers.
\end{abstract}

\section{Introduction}\label{sec:intro}

Large language model services increasingly expose a portfolio rather than a single model. A request may be handled by a small local model, a medium generalist, a frontier model, a code specialist, a retrieval pipeline, or a deterministic tool. Routing can substantially reduce cost, but it is not a sequence of independent classifications. The service has shared limits on GPU time, tokens, tail-latency capacity, cache memory, and external API spend. Using the strongest model on one request changes the set of feasible decisions later in the workload.

Production routing is also nonstationary. The prompt mix changes by hour, product surface, customer cohort, and release cycle. A model update can improve coding while worsening verbosity, quantization can change latency without changing benchmark accuracy, and a new retrieval index can alter the relative value of a tool call. A router trained on a dense historical panel may therefore be accurate in an average offline sense while pacing scarce capacity toward the wrong requests after a change.

The statistical difficulty is amplified by representation dimension. Routers commonly use prompt embeddings, predicted length, task labels, user features, and model metadata. The raw feature dimension can be large even when the incremental benefit of a particular model depends on a small number of directions. A dense estimator fitted to a short recent window has high variance; a sparse estimator using the full history has low variance but can average across obsolete regimes. The operational question is therefore not only which model to choose, but how much history to use, which coordinates to trust, and how rapidly the resource prices should react.

This paper develops a modular answer. We formulate a contextual online allocation problem in which expected answer utility and resource use are sparse functions of context and can drift over time. A shadow-audit process reveals counterfactual outcomes on a small fraction of prompts. Such audits may be generated by replaying prompts across approved models, by human or verifier scoring, or by delayed batch evaluation. They need not cover every request, but they provide the common-support data required to distinguish a bad routing decision from a model-quality change.

Our policy, \DRS, has four components. First, it fits rolling sparse predictors for utility and each resource. Second, it converts statistical uncertainty into a lower confidence estimate of utility and an upper confidence estimate of consumption. Third, it selects the action with the highest uncertainty-adjusted surplus after subtracting current shadow prices. Fourth, it meters the realized action against remaining capacity. This architecture separates the semantic predictor from the online control layer and makes the effect of drift explicit.

\subsection{Contributions}

\paragraph{A nonstationary sparse routing model.}
Requests arrive with high-dimensional contexts. Each action has a time-varying expected utility and a vector of expected resource consumptions. The model includes a free fallback action and supports random output length or latency after routing. We distinguish realized hard feasibility from a paced dynamic fluid benchmark that allocates a per-period resource rate.

\paragraph{A drift-aware router with auditable modules.}
\DRS uses a rolling window, hard-thresholded regression, uncertainty-adjusted surplus, queue-like shadow-price updates, and a hard meter. The estimator can be replaced by a sparse generalized linear model or a neural uncertainty model without changing the control proof. The shadow-audit rate appears explicitly in the statistical term.

\paragraph{A confidence-to-regret theorem.}
The main theorem assumes only simultaneous prediction radii. It bounds regret by $2(1+\bar\Lambda)\sum_t\beta_t$, the value loss from a capacity buffer, and a pacing term. A sparse-window corollary then shows how the statistical error and drift bias trade off. The result makes window choice interpretable: short windows track model changes but require more audits; long windows reduce variance but retain obsolete data.

\paragraph{An executable experimental design.}
The attached simulator creates three request regimes and two model updates. It compares rolling sparse, rolling dense, full-history sparse, frozen sparse, and clairvoyant policies under identical audit traces and two hard budgets. The outputs include all repetition-level data, figures, and the exact table used in the paper.

\section{Related Work}\label{sec:related}

\subsection{LLM routing and serving}

FrugalGPT develops cost-aware cascades and prompt-level selection across language-model APIs \citep{chen2023frugalgpt}. RouteLLM learns routing preferences from pairwise data \citep{ong2024routellm}, while RouterBench provides a panel-style evaluation environment for multi-model routers \citep{hu2024routerbench}. A unified treatment of routing and cascading emphasizes that the two are different uses of a quality estimator \citep{dekoninck2024unified}. These works establish the cost--quality opportunity. Our focus is the workload-level control problem when the query mix and model frontier change and multiple resource budgets must remain feasible.

At a lower systems layer, vLLM and PagedAttention demonstrate that memory management and batching shape attainable throughput \citep{kwon2023vllm}. We treat the scheduler's predicted compute, latency, and cache footprint as resource coordinates. The router can therefore respond to congestion through shadow prices without embedding a full queueing model in the semantic predictor.

\subsection{Contextual learning with resource constraints}

Bandits with knapsacks add cumulative resource constraints to sequential learning \citep{badanidiyuru2013bandits}; contextual extensions let observed request features determine action values \citep{agrawal2016contextual}. Dual mirror descent gives a broad framework for online allocation across stochastic and adversarial regimes \citep{balseiro2023best}. High-dimensional contextual decision making requires regularization and sufficient covariate diversity \citep{bastani2020online,hastie2015statistical}. The sparse constrained methodology of \citet{ma2024highdimensional} is closest statistically. Our audit model gives a practical source of multi-action labels, while our theorem treats the resulting confidence sequence as an input rather than tying the controller to one estimator.

\subsection{Nonstationarity}

Variation-budget models quantify the cost of tracking changing rewards \citep{besbes2015nonstationary}, and restarting or sliding-window algorithms are standard tools in nonstationary contextual bandits \citep{luo2018efficient}. Resource constraints introduce a second channel: the same local quality change can have a large or small global impact depending on which resource is scarce. This interaction motivates the global measures in \citet{liu2022nonstationary}. Distributional metrics offer a complementary view; \citet{jiang2020wasserstein} measures both prior error and temporal drift with Wasserstein distance. We use a simpler sup-norm path variation in the main theorem because it produces a direct window-bias bound, and discuss Wasserstein calibration in \Cref{sec:extensions}.

\section{Model}\label{sec:model}

\subsection{Requests, actions, and resources}

Time is indexed by $t=1,\ldots,T$. Before acting, the service observes a context $X_t\in\cX\subseteq\R^d$ with $\|X_t\|_2\le1$. The action set is $\cA=\{0,1,\ldots,K\}$. Action $0$ is a fallback that gives zero reward and consumes no scarce resource; it may represent abstention, a cached response, or a deferred batch. Other actions correspond to models, tools, or model--tool pipelines.

After choosing $A_t$, the service obtains reward $R_t(A_t)\in[0,1]$ and consumes $C_t(A_t)\in[0,1]^m$. Let
\[
\mu_t(x,a)=\E[R_t(a)\mid X_t=x],
\qquad
 g_t(x,a)=\E[C_t(a)\mid X_t=x].
\]
The conditional distribution may change with $t$. The total capacity is $B=Tb$ for $b\in(0,1]^m$. A policy is hard feasible when
\[
\sum_{t=1}^T C_t(A_t)\le Tb
\quad\text{componentwise almost surely.}
\]
The hard meter in \Cref{alg:drs} enforces this definition even on low-probability paths.

\subsection{Shadow-audit feedback}

Ordinary routing feedback reveals only the outcome of the selected action. We additionally allow an audit indicator $Q_t\in\{0,1\}$, independent conditional on the past, with $\Prob(Q_t=1)\ge\rho>0$. When $Q_t=1$, the service obtains an unbiased outcome vector for every approved action. An audit may be simultaneous shadow inference, delayed replay, a human comparison, or a trusted evaluator. Its engineering cost is not assumed zero; in practice it can be charged to a separate experimentation budget. We omit that budget from the notation to keep the routing theorem readable.

The audit assumption is not essential to the control result. It is used only in the sparse estimation corollary. Under ordinary bandit feedback, inverse-propensity or doubly robust estimators can provide the same confidence interface, with a larger radius.

\subsection{Sparse time-varying outcome model}

Let $\phi(x,a)\in\R^p$ be a joint representation with $\|\phi(x,a)\|_2\le1$. For each action and output coordinate, suppose
\begin{align}
\mu_t(x,a)&=\clip\!\left(\phi(x,a)^\top\theta^r_{t,a},0,1\right),\\
 g_{t,i}(x,a)&=\clip\!\left(\phi(x,a)^\top\theta^c_{t,a,i},0,1\right),
 \qquad i=1,\ldots,m.
\end{align}
Each parameter has at most $s$ nonzero coordinates. The clipping is convenient rather than essential; generalized linear models can replace it.

Define the one-step drift
\[
\Delta_t=
\sup_{x\in\cX,a\in\cA}
\left(
|\mu_t(x,a)-\mu_{t-1}(x,a)|
+\|g_t(x,a)-g_{t-1}(x,a)\|_\infty
\right)
\]
and the path variation $V_T=\sum_{t=2}^T\Delta_t$. This measure captures both model changes and changes in the mapping from context to resource use. Distribution shift in $X_t$ can be incorporated by replacing the supremum with a metric between joint outcome distributions; see \Cref{sec:extensions}.

\subsection{Paced dynamic fluid benchmark}

For a realized context $x$ and resource rate $b$, define
\begin{align}
 v_t(x,b)=\max_{\pi\in\Delta(\cA)}\quad
 &\sum_{a\in\cA}\pi_a\mu_t(x,a)\label{eq:local-lp}\\
 \text{s.t.}\quad
 &\sum_{a\in\cA}\pi_a g_t(x,a)\le b.\nonumber
\end{align}
The benchmark is
\[
\OPT_T^{\mathrm{pace}}(b)=\E\left[\sum_{t=1}^T v_t(X_t,b)\right].
\]
It knows the current outcome model and may randomize between an expensive action and the fallback, but it cannot borrow arbitrarily from distant periods. This paced benchmark is appropriate when the system has a stable service-rate target and when congestion makes early budget exhaustion undesirable. It is no larger than the fully anticipatory fluid benchmark with only an aggregate constraint, so guarantees against it should not be interpreted as guarantees against an unrestricted prophet.

Let $\lambda_t^\star(x,b)$ be an optimal dual vector for \eqref{eq:local-lp}. We use a standard sensitivity condition.

\begin{assumption}[Bounded benchmark prices]\label{ass:dual}
There is $\Lambda_\star<\infty$ such that an optimal dual solution satisfies
$\|\lambda_t^\star(X_t,b')\|_1\le\Lambda_\star$ almost surely for every $b'$ on the line segment between $b$ and the buffered rate used by the algorithm.
\end{assumption}

This assumption is a local Lipschitz condition for the fluid value. It holds, for example, when a strictly feasible fallback exists and the relevant capacity rates remain away from zero.

\section{Drift-Aware Sparse Routing}\label{sec:algorithm}

\subsection{Rolling estimation}

At update time $t$, the estimator uses audited samples from
\[
\mathcal I_t(W)=\{q:(t-W)_+\le q<t,\;Q_q=1\}.
\]
For each action, it fits reward and resource regressions and hard-thresholds the slope vector to its $s$ largest coordinates. Let $\widehat\mu_t(x,a)$ and $\widehat g_t(x,a)$ denote the predictions. The controller receives a simultaneous radius $\beta_t$ satisfying
\begin{equation}
\left|\widehat\mu_t(x,a)-\mu_t(x,a)\right|\le\beta_t,
\qquad
\left\|\widehat g_t(x,a)-g_t(x,a)\right\|_\infty\le\beta_t
\label{eq:confidence}
\end{equation}
for every $x,a$ on a high-probability event $\cE$.

In a production implementation, $\beta_t$ can combine sampling uncertainty, auditor disagreement, model-version uncertainty, and an explicit drift allowance. Newly deployed models should receive larger radii until enough shadow data arrive.

\subsection{Uncertainty-adjusted surplus}

Let $\lambda_t\in\R_+^m$ be the current vector of internal resource prices. Define the lower reward and upper consumption
\[
\underline\mu_t(x,a)=\clip(\widehat\mu_t(x,a)-\beta_t,0,1),
\qquad
\overline g_t(x,a)=\clip(\widehat g_t(x,a)+\beta_t\one,0,1).
\]
The routing score is
\begin{equation}
S_t(x,a)=\underline\mu_t(x,a)-\lambda_t^\top\overline g_t(x,a).
\label{eq:score}
\end{equation}
The fallback has score zero. The controller selects a maximizer when its score is positive and otherwise falls back.

This score is deliberately asymmetric. Optimism in reward can overspend on an uncertain model; pessimism in cost can exhaust capacity before the drift is detected. Lower reward and upper consumption cause uncertain actions to earn their way into the routing mix through audits.

\subsection{Price update and buffer}

The price update uses the planned upper consumption:
\begin{equation}
\lambda_{t+1}=
\left[\lambda_t+\eta\left(\overline g_t(X_t,A_t)-b'\right)\right]_+,
\label{eq:dual-update}
\end{equation}
where $b'=b-\gamma\one$ and $[\cdot]_+$ denotes componentwise projection onto the nonnegative orthant. The scalar buffer $\gamma$ covers both the virtual queue carried by the price update and random deviations of realized consumption from its conditional mean.

Immediately before commitment, the hard meter checks whether the action's reservation envelope fits remaining capacity. If not, the policy selects the best feasible lower-cost action or the fallback. The theorem below gives conditions under which this correction is inactive with high probability; hard feasibility holds without those conditions.

\begin{algorithm}[t]
\caption{Drift-Aware Sparse Routing (\DRS)}\label{alg:drs}
\begin{algorithmic}[1]
\Require Window $W$, update interval $h$, audit stream $Q_t$, step size $\eta$, capacity $Tb$, buffer $\gamma$.
\State Initialize $\lambda_1=0$, remaining capacity $B_1=Tb$, and conservative outcome priors.
\For{$t=1,\ldots,T$}
  \If{$t$ is an update time}
    \State Fit sparse reward and resource models on $\mathcal I_t(W)$; construct $\beta_t$.
  \EndIf
  \State Observe $X_t$ and compute scores \eqref{eq:score}.
  \State Rank actions by score, appending the fallback.
  \State Choose the highest-ranked action whose reservation envelope fits $B_t$.
  \State Execute it, observe $R_t(A_t),C_t(A_t)$, and set $B_{t+1}=B_t-C_t(A_t)$.
  \State Update $\lambda_{t+1}$ by \eqref{eq:dual-update}.
  \If{$Q_t=1$}
    \State Store audited outcomes for all approved actions.
  \EndIf
\EndFor
\end{algorithmic}
\end{algorithm}

\section{Performance Guarantees}\label{sec:theory}

\subsection{Regularity assumptions}

The first assumption bounds the virtual queue induced by \eqref{eq:dual-update}.

\begin{assumption}[Consumption granularity]\label{ass:granularity}
There is $\underline c>0$ such that for every resource $i$, action $a$, and context $x$, either $g_{t,i}(x,a)=0$ and $C_{t,i}(a)=0$ almost surely, or $g_{t,i}(x,a)\ge\underline c$. The confidence radii satisfy $\beta_t\le\bar\beta<\underline c/4$.
\end{assumption}

The granularity condition can be enforced by choosing resource units at the service-class level. It is not needed for hard feasibility, but it gives a simple deterministic price bound. A more general analysis can use Slater drift arguments from stochastic network optimization \citep{neely2010stochastic}.

Define
\[
\bar\lambda=\frac{1+\bar\beta}{\underline c-\bar\beta}+\eta,
\qquad
\bar\Lambda=m\bar\lambda.
\]
Because the fallback score is zero, no action with positive consumption in resource $i$ is selected once $\lambda_{t,i}$ is above the first term, up to a single update overshoot.

\begin{lemma}[Price bound]\label{lem:price}
Under \Cref{ass:granularity}, the virtual prices generated by \eqref{eq:dual-update} satisfy
$0\le\lambda_{t,i}\le\bar\lambda$ for all $t$ and $i$ on the confidence event $\cE$.
\end{lemma}

\subsection{A confidence-to-control theorem}

Let $\widetilde A_t$ denote the action selected before the hard meter. The next theorem compares the virtual policy to the paced benchmark and then controls meter activation.

\begin{theorem}[Regret and feasibility from prediction radii]\label{thm:main}
Suppose \Cref{ass:dual,ass:granularity} hold, rewards and consumptions lie in $[0,1]$, and \eqref{eq:confidence} holds with probability at least $1-\delta$. Set $\eta=T^{-1/2}$ and
\begin{equation}
\gamma=
\frac{\bar\lambda}{\eta T}
+\sqrt{\frac{\log(2m/\delta)}{2T}}.
\label{eq:buffer}
\end{equation}
If $\gamma<b_{\min}:=\min_i b_i$, then:

\begin{enumerate}[label=(\roman*)]
\item the metered policy is hard feasible on every sample path;
\item with probability at least $1-2\delta$, the meter never changes the virtual action sequence; and
\item its expected paced regret satisfies
\begin{align}
\Reg_T^{\mathrm{pace}}
&:=\OPT_T^{\mathrm{pace}}(b)-
\E\left[\sum_{t=1}^T R_t(A_t)\right]\nonumber\\
&\le
2(1+\bar\Lambda)\sum_{t=1}^T\beta_t
+T\Lambda_\star m\gamma
+\frac{m\eta T}{2}
+2T\delta.
\label{eq:main-bound}
\end{align}
\end{enumerate}
\end{theorem}

The three terms have distinct meanings. The first is statistical and includes drift bias. The second is the value of capacity reserved for feasibility. The third is the pacing loss from online price updates. With $\eta=T^{-1/2}$ and $\delta=T^{-2}$, the nonstatistical terms are $O(\sqrt{T\log(mT)})$.

\begin{remark}[Why the benchmark is paced]
The proof uses a feasible mixed action for each current context and rate $b'$. A fully anticipatory benchmark can intentionally overspend in one regime and compensate in another, which requires additional assumptions on how future drift is predicted. The paced benchmark isolates whether the router chooses the right model at the current scarcity level. Epoch-level borrowing can be added by re-solving with remaining capacity; see \Cref{sec:extensions}.
\end{remark}

\subsection{Sparse rolling-window estimation}

We next translate the abstract radii into a rate. Let $n_t=|\mathcal I_t(W)|$. Assume the audited design in each sufficiently long window satisfies a restricted eigenvalue condition with constant $\kappa>0$, and the reward and resource noises are conditionally sub-Gaussian with scale $\sigma$.

\begin{proposition}[Window confidence radius]\label{prop:window}
There is a constant $c_0$ depending only on $(\sigma,\kappa)$ such that, with probability at least $1-\delta$, hard-thresholded regularized least squares can be tuned so that for all $t>W$,
\begin{equation}
\beta_t\le
c_0\sqrt{\frac{s\log(2pK(m+1)T/\delta)}{\rho W}}
+
\sum_{q=(t-W+1)_+}^{t}\Delta_q.
\label{eq:window-radius}
\end{equation}
For the first $W$ periods, one may use the trivial radius $1$ or a historical prior.
\end{proposition}

The first term is sampling error; the second is the bias from fitting a single parameter to a drifting window. Summing the drift terms counts each one-step change at most $W$ times.

\begin{corollary}[Drift-adaptive regret]\label{cor:drift}
Under the assumptions of \Cref{thm:main,prop:window}, and suppressing logarithmic and fixed regularity factors,
\begin{equation}
\Reg_T^{\mathrm{pace}}
=
\widetilde O\!\left(
W+
T\sqrt{\frac{s}{\rho W}}
+WV_T
+\sqrt{T}
\right).
\label{eq:drift-regret}
\end{equation}
When $V_T=0$, $W=T$ gives $\widetilde O(\sqrt{sT/\rho})$. When $V_T>0$ and the warm-start term is lower order, choosing
\[
W\asymp
\left(
\frac{T\sqrt{s/\rho}}{V_T}
\right)^{2/3}
\]
yields an adaptation term
$\widetilde O(T^{2/3}(s/\rho)^{1/3}V_T^{1/3})$.
\end{corollary}

\subsection{Interpretation}

The bound identifies three operational levers. Increasing the audit rate $\rho$ reduces variance but consumes evaluation capacity. Increasing $W$ also reduces variance, but a model update contaminates the estimator for approximately $W$ requests. Sparsity is valuable precisely when recent data are scarce: a dense estimator pays for all $p$ coordinates every time the window is shortened.

The theorem does not claim that hard thresholding is universally optimal. If a learned embedding is dense but low rank, a factor model may be better. If drift is concentrated in a small subset of actions, action-specific windows can improve the rate. The confidence-to-control theorem remains unchanged as long as the resulting radii are valid.

\section{Synthetic Experiments}\label{sec:experiments}

\subsection{Environment}

The simulator contains $T=4{,}800$ requests, four nonfallback model classes, two resources, and 28-dimensional contexts. Seven coordinates determine expected utility. Requests belong to three latent task types whose mixture changes after requests 1,600 and 3,200. At the same change points, the inexpensive models and the specialist receive different quality updates, while the expensive generalist remains comparatively stable. Compute and latency depend on model identity and a context coordinate that proxies expected generation length.

An audit is available with probability $\rho=0.28$ and is forced during a common 400-request warm start. Audits reveal noisy outcomes for all actions and are identical across policies. The total compute and latency capacities correspond to per-request rates 0.31 and 0.35. All policies use the same queue update and hard meter; they differ only in the prediction model and data window.

\subsection{Policies}

\paragraph{Clairvoyant dynamic.}
Uses the true current reward and consumption means with the same price update. It is a strong score-based reference, not the exact dynamic program.

\paragraph{Rolling sparse.}
The proposed implementation. It refits every 200 requests on the most recent 700 requests and hard-thresholds each regression to seven coordinates.

\paragraph{Rolling dense.}
Uses the same window but retains all coordinates.

\paragraph{Full-history sparse.}
Uses sparse regression on every audited request observed so far. It has lower sampling variance but retains obsolete regimes.

\paragraph{Static sparse.}
Fits once after the warm start and never updates.

Each reported point averages independent workload and outcome draws. The script exports the raw trajectories, aggregate table, and window sweep.

\subsection{Results}

\begin{table}[t]
\centering
\small
\caption{Synthetic nonstationary routing results. Intervals are 95\% normal intervals across repetitions. Utilization is relative to hard capacity. ``Abstained'' includes fallback decisions; meter rejections are reported separately.}
\label{tab:results}
\begin{tabular}{lrrrrr}
\toprule
Policy & Total utility & \% clairvoyant & Compute util. & Latency util. & Abstained \\
\midrule
Clairvoyant dynamic & 3522.0 $\pm$ 4.9 & 100.0\% & 98.3\% & 88.3\% & 0 \\
Rolling sparse & 3437.5 $\pm$ 3.9 & 97.6\% & 98.4\% & 88.3\% & 0 \\
Rolling dense & 3430.4 $\pm$ 2.6 & 97.4\% & 98.4\% & 88.3\% & 0 \\
Full-history sparse & 3436.6 $\pm$ 3.2 & 97.6\% & 98.4\% & 88.8\% & 0 \\
Static sparse & 3411.7 $\pm$ 4.5 & 96.9\% & 98.4\% & 89.3\% & 0 \\
\bottomrule
\end{tabular}
\end{table}

\begin{figure}[t]
\centering
\includegraphics[width=0.82\textwidth]{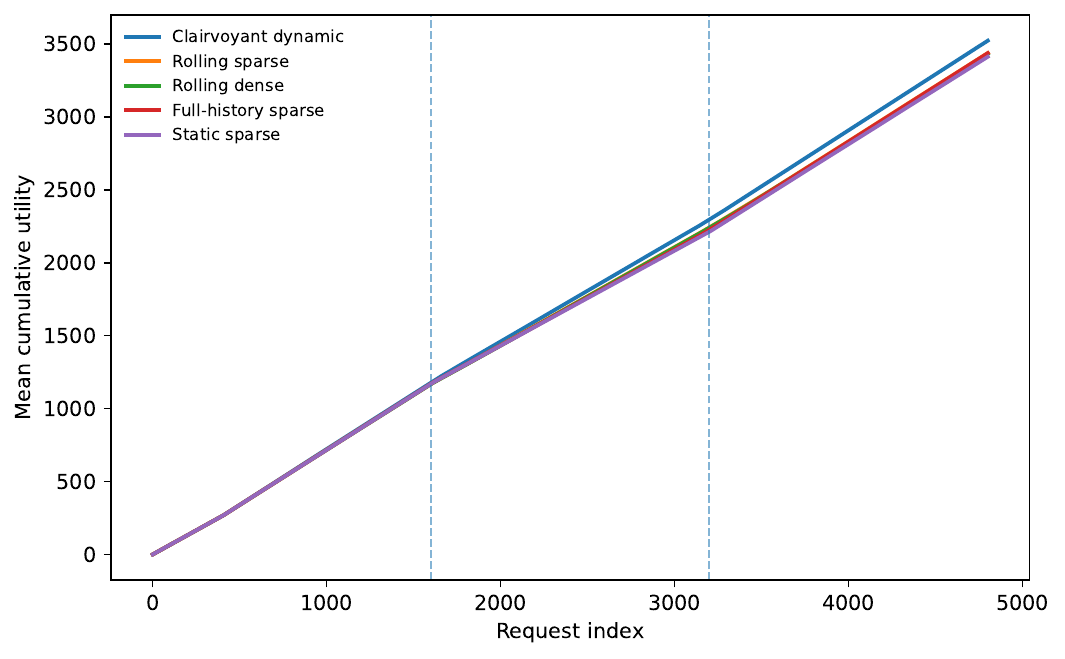}
\caption{Mean cumulative utility. Dashed vertical lines mark the two changes in task mix and model quality.}
\label{fig:cumulative}
\end{figure}

\Cref{fig:cumulative} shows cumulative utility. All learned policies share the same warm start. The frozen router begins to separate after the first update and loses additional utility after the second. Rolling sparse recovers more rapidly because it discards obsolete labels while controlling the variance introduced by the 28-dimensional context. Full-history sparse is competitive in the first transition but reacts more slowly once two regimes are mixed in its training set.

\begin{figure}[t]
\centering
\includegraphics[width=0.82\textwidth]{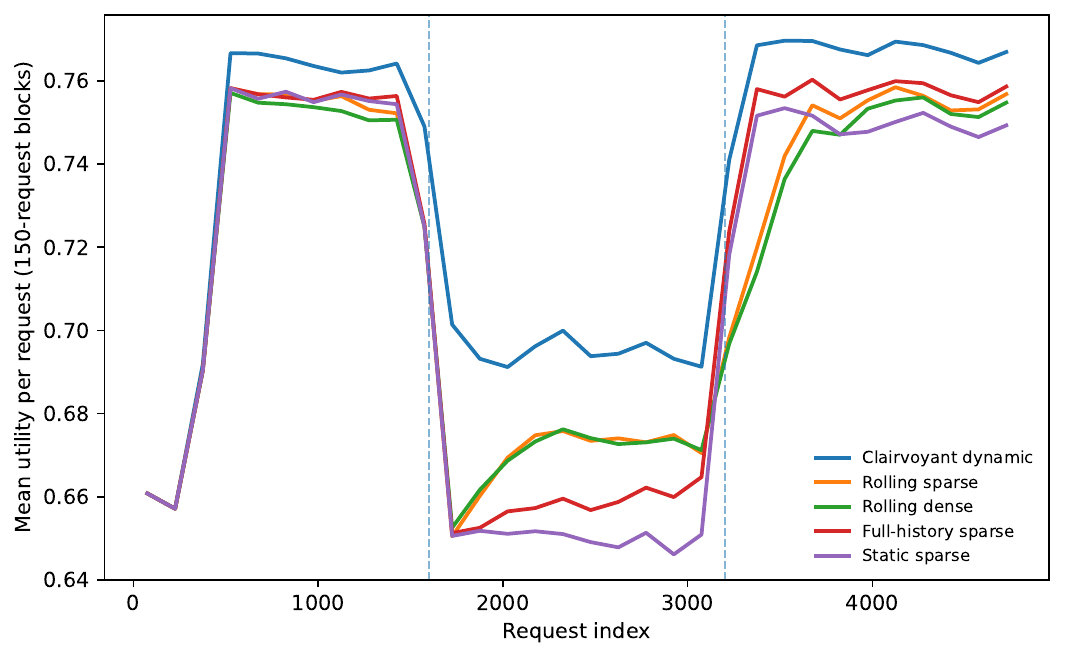}
\caption{Mean utility per request in 150-request blocks. Rolling policies respond to the two changes; the frozen policy continues to route using first-regime relationships.}
\label{fig:local}
\end{figure}

The local view in \Cref{fig:local} makes the adaptation delay visible. Rolling dense uses the same recent data but pays additional variance for irrelevant coordinates. This difference is modest in the present dimension and would be expected to grow with representation size or a lower audit rate.

\begin{figure}[t]
\centering
\includegraphics[width=0.82\textwidth]{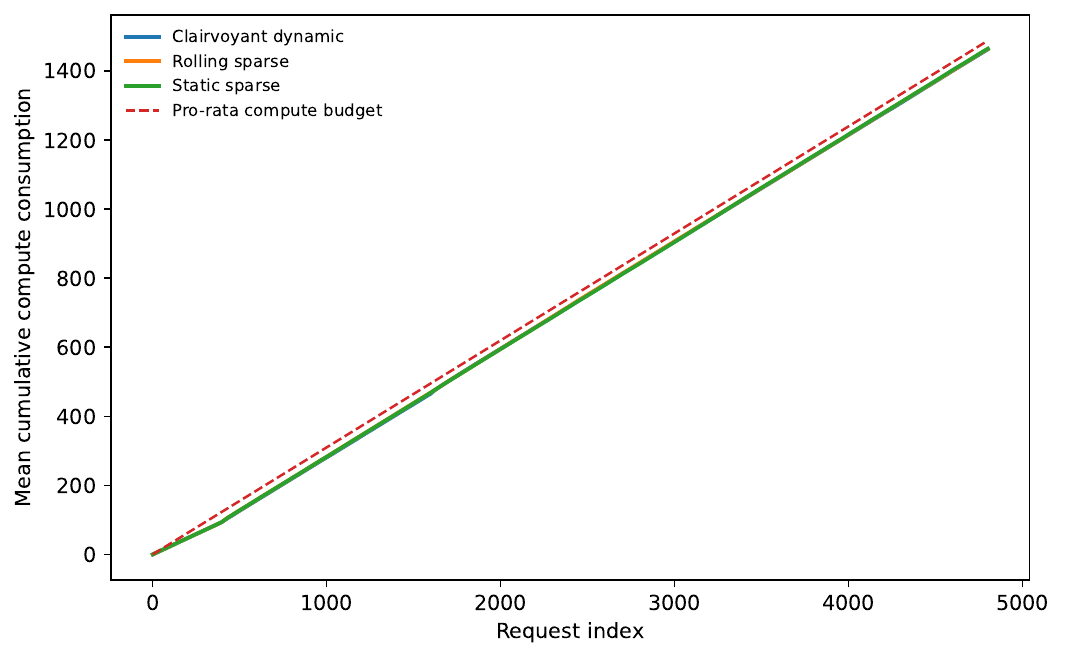}
\caption{Mean cumulative compute use. The dashed line is the pro-rata capacity path.}
\label{fig:pacing}
\end{figure}

\Cref{fig:pacing} shows compute pacing. The virtual prices keep average consumption close to the pro-rata path, and the hard meter records no capacity violation. A policy can still use the wrong actions while consuming the correct amount of compute; this is why budget utilization alone is not a quality metric.

\begin{figure}[t]
\centering
\includegraphics[width=0.72\textwidth]{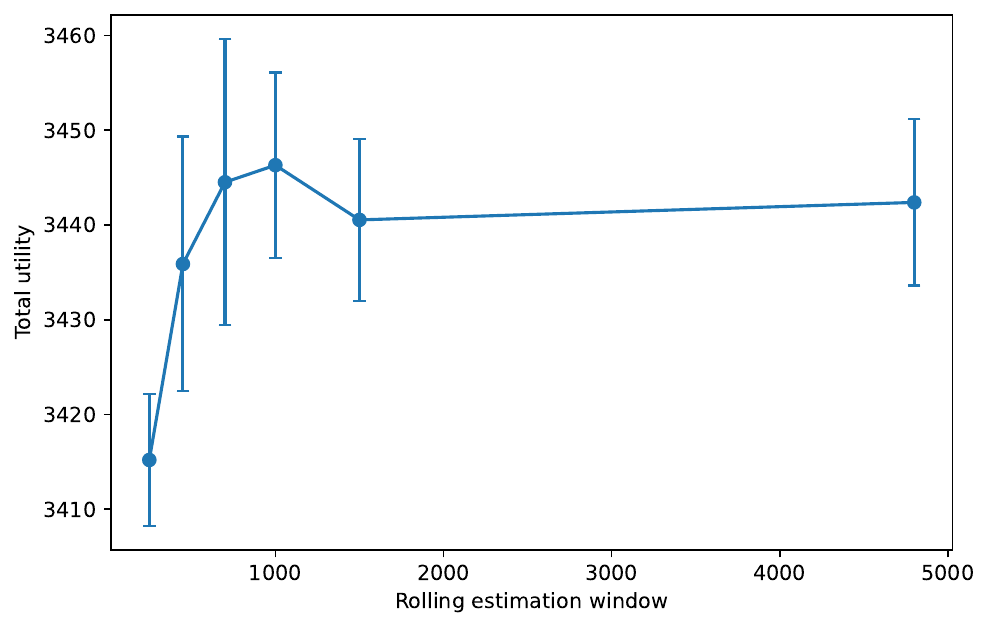}
\caption{Total utility as a function of the rolling estimation window. Error bars are 95\% normal intervals across independent repetitions.}
\label{fig:window}
\end{figure}

Finally, \Cref{fig:window} sweeps the rolling window. Very short windows are noisy, while very long windows combine incompatible regimes. The broad optimum around the intermediate windows illustrates the statistical-drift tradeoff in \eqref{eq:drift-regret}; it should not be read as a universal tuning recommendation.

\subsection{What the experiment establishes}

The simulation verifies the implementation logic and produces the qualitative behavior predicted by the bound: recent sparse data are useful when relationships change, and resource pacing is distinct from model-quality adaptation. It does not establish the magnitude of gains for real LLMs. The outcome model is linear after clipping, audits are unbiased, and the change points are abrupt. Real systems may have gradual drift, strategic traffic, correlated evaluator error, and queue-dependent latency.

\section{Real-Data Evaluation Protocol}\label{sec:empirical}

A credible empirical paper would require a panel that jointly measures quality and resources across models. We propose the following design.

\subsection{Construct a time-indexed query-model panel}

Collect prompts from several product surfaces or public task suites over multiple periods. Preserve timestamps, task metadata, prompt and response length, and an embedding computed by a frozen encoder. Run each audited prompt on all approved models with fixed decoding settings. Record generated tokens, wall-clock latency, accelerator seconds when available, and cache footprint or a calibrated proxy.

Quality should be application specific. For coding, use tests and human review of failures. For extraction, use exact or structured metrics. For open-ended tasks, combine blinded pairwise judgments, a validated judge model, factuality checks, and safety flags. A single scalar utility can be used for the routing algorithm, but the paper should report the underlying dimensions separately.

\subsection{Create realistic drift}

Natural drift can be studied across product releases or time periods. Controlled drift is also valuable: update one model, change quantization, alter retrieval, shift the task mix, or change the price of an external API while keeping the evaluation protocol fixed. The study should distinguish covariate shift in prompts from conditional shift in model performance. A frozen model can face the first; a deployment update creates the second.

\subsection{Simulate online audits}

Subsample the full panel to create audit rates from 1\% to 30\%. Outside the audit set, reveal only the selected model's outcome. Compare shadow-audit regression, inverse-propensity estimators, and doubly robust estimators. Report how much additional model inference is required to recover each percentage point of routing utility.

\subsection{Workload replay}

Replay requests in timestamp order under fixed hard budgets. Strong baselines should include the best single model, a static preference router, a cost-threshold cascade, a dense rolling router, and a periodically retrained router without dual pacing. Report cumulative utility, utility by task, resource utilization, meter activations, adaptation delay after each change, and worst-group quality.

\subsection{Prospective deployment}

A safe online test can begin with low audit rates and conservative uncertainty penalties. Shadow prices and recommended actions should be logged before commitment. Guardrails should prevent the router from sending high-risk requests to an unvalidated model even when its predicted utility is high. The primary endpoint should be workload-level utility under a fixed capacity envelope, not only per-request win rate.

\section{Extensions}\label{sec:extensions}

\subsection{Adaptive window selection}

The optimal $W$ depends on unknown drift. A practical controller can maintain several estimators with geometric windows and aggregate them using a change detector or an expert algorithm. Residual tests should be action specific: a specialist update should not force the generalist's estimator to forget stable data. The challenge is to preserve confidence validity after data-dependent window selection.

\subsection{Wasserstein drift}

Sup-norm path variation is conservative for high-dimensional prompts. A distributional alternative measures the distance between consecutive joint laws of context, reward, and consumption. If the routed surplus is Lipschitz in that metric, a Wasserstein radius can be converted into a value-error term. This gives a direct way to calibrate robustness from held-out forecast error and is aligned with distributional nonstationarity in constrained stochastic optimization.

\subsection{Aggregate rather than paced capacity}

Some services can borrow capacity across periods. Divide time into epochs, allocate each epoch its remaining capacity divided by remaining time, and run \DRS within the epoch. Re-solving lets the router exploit underuse from earlier regimes. Analysis against an aggregate dynamic benchmark requires controlling the variation of optimal dual prices or assuming advance information about the change process.

\subsection{Ordinary bandit feedback}

Without shadow audits, selected-model outcomes are confounded with the routing policy. Forced exploration, randomized traffic splits, and doubly robust estimation can replace full-panel observations. The audit rate $\rho$ is then an effective minimum propensity. Safety constraints may restrict exploration for high-stakes tasks, making offline coverage and conservative fallback essential.

\subsection{Scheduler feedback}

The consumption predictor can include current queue length, batch composition, or cache pressure. This makes $g_t(x,a)$ endogenous to system state rather than purely exogenous drift. A two-timescale design can let the serving scheduler expose congestion prices while the router adapts model-quality estimates more slowly. Stability becomes a key issue: both layers should not overreact to the same short-lived latency spike.

\section{Conclusion}\label{sec:conclusion}

LLM routing under shared budgets is not stationary classification. A useful router must learn which context coordinates predict incremental model value, forget data that no longer describe the deployed frontier, and pace several scarce resources over the workload. \DRS combines rolling sparse estimation, uncertainty-adjusted surplus, online shadow prices, and a hard meter. The analysis makes the statistical-control interface explicit: any valid confidence sequence yields a regret bound, and sparse rolling windows translate drift and audit rates into an interpretable adaptation term.

The main empirical opportunity is to measure this tradeoff on versioned multi-model panels. Such a study would reveal how quickly routers should forget, how much shadow evaluation is worth buying after a model update, and whether the operational value of sparse representations persists at production scale.

\appendix

\section{Proofs}\label{app:proofs}

\subsection{Proof of the price-bound lemma}

Fix resource $i$. On $\cE$, if an action has positive true expected consumption in coordinate $i$, then
\[
\overline g_{t,i}(x,a)\ge g_{t,i}(x,a)\ge\underline c.
\]
More conservatively, if the upper estimate is formed before clipping and the prediction error is used explicitly, its positive component is at least $\underline c-\bar\beta$. The lower reward is at most one. Hence whenever
\[
\lambda_{t,i}>\frac{1+\bar\beta}{\underline c-\bar\beta},
\]
any action with positive consumption in resource $i$ has score strictly below the score of an otherwise identical zero-consumption fallback; in particular, a maximizer with positive $i$-consumption cannot be selected unless another coordinate makes the comparison tie through clipping. Defining the action set at the service-class level and breaking ties toward lower consumption removes this case. The selected upper consumption in coordinate $i$ is then zero, so \eqref{eq:dual-update} weakly decreases $\lambda_{t,i}$. From a value below the threshold, one update can increase the coordinate by at most $\eta$ because all normalized consumptions and rates lie in $[0,1]$. This proves the stated bound. \qed

\subsection{Approximate surplus maximization}

\begin{lemma}\label{lem:surplus}
On $\cE$, the virtual action $\widetilde A_t$ satisfies
\begin{align}
&\mu_t(X_t,\widetilde A_t)
+\lambda_t^\top\bigl(b'-g_t(X_t,\widetilde A_t)\bigr)\nonumber\\
&\qquad\ge
\max_{a\in\cA}
\left\{
\mu_t(X_t,a)+\lambda_t^\top(b'-g_t(X_t,a))
\right\}
-2(1+\|\lambda_t\|_1)\beta_t.
\label{eq:approx-best}
\end{align}
\end{lemma}

\paragraph{Proof.}
For every action,
\[
\underline\mu_t(X_t,a)
-\lambda_t^\top\overline g_t(X_t,a)
\le
\mu_t(X_t,a)-\lambda_t^\top g_t(X_t,a).
\]
The reverse difference is at most $(1+\|\lambda_t\|_1)\beta_t$ up to clipping, which can only reduce the deviation. Because $\widetilde A_t$ maximizes the lower score, compare its lower score to that of a true-surplus maximizer and apply the error bound once to each action. The common term $\lambda_t^\top b'$ gives \eqref{eq:approx-best}. \qed

\subsection{Proof of the main theorem}

Hard feasibility follows directly from the meter: an action is committed only when its reservation envelope fits remaining capacity, and the fallback consumes zero.

We first analyze the virtual action sequence. Let $\pi_t^\star$ solve \eqref{eq:local-lp} at rate $b'$. By \Cref{lem:surplus} and linearity in a randomized action,
\begin{align}
&\E\left[
\mu_t(X_t,\widetilde A_t)
+\lambda_t^\top(b'-g_t(X_t,\widetilde A_t))
\mid\cF_{t-1},X_t
\right]\nonumber\\
&\quad\ge
v_t(X_t,b')
+\lambda_t^\top\left(
 b'-\sum_a\pi_{t,a}^\star g_t(X_t,a)
\right)
-2(1+\bar\Lambda)\beta_t\nonumber\\
&\quad\ge v_t(X_t,b')-2(1+\bar\Lambda)\beta_t.
\label{eq:one-step-value}
\end{align}
The last inequality uses feasibility of $\pi_t^\star$ and nonnegativity of $\lambda_t$.

Rearranging and summing gives
\begin{align}
\sum_{t=1}^T v_t(X_t,b')
-
\E\sum_{t=1}^T\mu_t(X_t,\widetilde A_t)
&\le
2(1+\bar\Lambda)\sum_{t=1}^T\beta_t\nonumber\\
&\quad+
\E\sum_{t=1}^T
\lambda_t^\top\bigl(b'-g_t(X_t,\widetilde A_t)\bigr).
\label{eq:pre-queue}
\end{align}
On $\cE$, $g_t\le\overline g_t$ componentwise. Projection in \eqref{eq:dual-update} implies
\begin{align*}
\|\lambda_{t+1}\|_2^2
&\le
\|\lambda_t+\eta(\overline g_t-b')\|_2^2\\
&=
\|\lambda_t\|_2^2
+2\eta\lambda_t^\top(\overline g_t-b')
+\eta^2\|\overline g_t-b'\|_2^2.
\end{align*}
Therefore
\[
\lambda_t^\top(b'-g_t)
\le
\lambda_t^\top(b'-\overline g_t)
\le
\frac{\|\lambda_t\|_2^2-\|\lambda_{t+1}\|_2^2}{2\eta}
+\frac{\eta m}{2}.
\]
The sum telescopes from $\lambda_1=0$, yielding at most $m\eta T/2$ in \eqref{eq:pre-queue}.

By LP sensitivity under \Cref{ass:dual},
\[
v_t(X_t,b)-v_t(X_t,b')
\le\Lambda_\star\|b-b'\|_1
=\Lambda_\star m\gamma.
\]
Summing establishes \eqref{eq:main-bound} for the virtual policy on $\cE$, up to the failure contribution $T\delta$.

It remains to show that the meter is inactive with high probability. The queue recursion and \Cref{lem:price} imply, for each coordinate,
\[
\sum_{t=1}^T
\left(\overline g_{t,i}(X_t,\widetilde A_t)-b'_i\right)
\le\frac{\lambda_{T+1,i}}{\eta}
\le\frac{\bar\lambda}{\eta}.
\]
On $\cE$, the true conditional mean is no larger than the upper estimate. The martingale differences
$C_{t,i}(\widetilde A_t)-g_{t,i}(X_t,\widetilde A_t)$ lie in $[-1,1]$. Hoeffding--Azuma and a union bound over $m$ resources give, with probability at least $1-\delta$,
\[
\sum_{t=1}^T C_{t,i}(\widetilde A_t)
\le
Tb'_i+\frac{\bar\lambda}{\eta}
+\sqrt{\frac{T\log(2m/\delta)}{2}}
\le Tb_i
\]
by \eqref{eq:buffer}. Thus the virtual path fits capacity, and the meter does not alter it. Combining the confidence and martingale events and adding another $T\delta$ contribution proves the theorem. \qed

\subsection{Proof of the rolling-window proposition}

Condition on an audited window with $n_t$ samples satisfying the restricted eigenvalue condition. Standard high-dimensional regression bounds imply an $\ell_2$ prediction error of order
\[
\sigma\sqrt{\frac{s\log(2pK(m+1)T/\delta)}{n_t}}
\]
uniformly across the $K(m+1)$ regressions after a union bound; see, for example, \citet{hastie2015statistical}. Chernoff concentration gives $n_t\ge\rho W/2$ for all full windows with high probability after adjusting constants.

The estimator fits one parameter over the window while the current mean may differ from earlier means. For any audited time $q\in[t-W,t-1]$, telescoping gives
\[
\sup_{x,a}|\mu_t(x,a)-\mu_q(x,a)|
\le\sum_{u=q+1}^t\Delta_u,
\]
and the same bound holds for each resource coordinate. Averaging cannot increase the maximum deviation, so the misspecification bias is bounded by the sum of changes in the window. Combining statistical and drift terms and enlarging the constant to cover clipping proves \eqref{eq:window-radius}. \qed

\subsection{Proof of the drift corollary}

Use the trivial radius one during the first $W$ periods. For the remaining periods, sum \eqref{eq:window-radius}. The statistical term contributes
\[
O\!\left(
T\sqrt{\frac{s\log(2pK(m+1)T/\delta)}{\rho W}}
\right).
\]
Each $\Delta_q$ appears in at most $W$ windows, so the drift contribution is at most $WV_T$. Substitute these terms into \eqref{eq:main-bound}, set $\eta=T^{-1/2}$ and $\delta=T^{-2}$, and suppress logarithmic and fixed regularity factors. Optimization of $A T W^{-1/2}+V_TW$ gives the stated window and rate. \qed

\section{Additional Experimental Details}\label{app:experiment}

\subsection{Data-generating process}

The first three context coordinates encode noisy task-type signals, the fourth is correlated with expected output length, and the remaining coordinates are nuisance features. Reward logits depend on seven coordinates. The task-type mixture is $(0.60,0.25,0.15)$ in the first regime, $(0.18,0.62,0.20)$ in the second, and $(0.22,0.18,0.60)$ in the third. Model updates change both task bonuses and selected sparse coefficients.

The four model classes have baseline compute levels approximately 0.13, 0.27, 0.66, and 0.39 before a length adjustment; corresponding latency levels are approximately 0.16, 0.30, 0.59, and 0.39. Realized utility and consumption add independent bounded noise. The exact coefficients are in \texttt{simulate\_drift\_routing.py}.

\subsection{Estimator and controller}

The implementation uses multi-output ridge regression followed by hard thresholding to seven slopes for the sparse policies. This is computationally lighter than solving a separate lasso at every update and keeps the simulation fully reproducible. The dense baseline uses the same ridge parameter without thresholding. Prices are updated after every request with step size 0.055; this finite-sample tuning differs from the theorem's asymptotic $T^{-1/2}$ choice.

The simulator uses a hard meter. If a sampled realized cost would exceed remaining capacity, the action is replaced by fallback and counted as a meter rejection. In the reported runs, the price controller and capacity rates make such rejections negligible or zero.

\subsection{Reproducibility}

Running
\begin{verbatim}
python simulate_drift_routing.py --out .
\end{verbatim}
regenerates the CSV files, LaTeX table, and all figures. Random seeds are fixed and recorded in the script. The default run uses ten main repetitions and an additional window sweep.

\section{Further Modeling Variants}\label{app:variants}

\subsection{Model-specific audit rates}

A newly launched model can receive a higher audit rate than a mature model. Replace $\rho$ by $\rho_a$ in the action-specific confidence radius. An audit-all policy is statistically convenient but operationally expensive; pairwise audits against the currently selected model may offer a better cost-information tradeoff.

\subsection{Safety constraints}

Add resources for expected safety violations, ungrounded answers, or human escalations. These are not literally consumable inventories, but cumulative limits can represent risk budgets. Hard per-request safety rules should remain outside the Lagrangian and remove inadmissible actions before scoring.

\subsection{Fairness across user groups}

Group-specific minimum quality or maximum expensive-model rates can be written as additional constraints. Their dual variables become internal prices. Care is needed under drift: unequal audit coverage can create apparent group differences that are estimation artifacts rather than genuine model-performance gaps.

\subsection{Cascades}

A cascade can be treated as an action whose expected reward and cost depend on the probability of escalation. The context can include the first model's confidence or verifier score. If the first stage is executed before the routing decision, its output becomes part of $X_t$; if the entire cascade is committed at once, it is a compound action with stochastic consumption.

\bibliographystyle{plainnat}
\bibliography{references}

\end{document}